\documentclass[conference]{IEEEtran}

\usepackage[cmex10]{amsmath}
\usepackage{amsthm}
\usepackage{amssymb}
\usepackage{mathrsfs}
\usepackage{graphicx}
\usepackage{float}
\usepackage{array}
\usepackage{epstopdf}
\usepackage{multirow}
\usepackage{color,soul}
\usepackage{epstopdf}
\usepackage{subcaption}

\begin{document}


\title{\vspace{0.5cm}MTBI Identification From Diffusion MR Images Using Bag of Adversarial Visual  Features}

\author{Shervin~Minaee$^{1}$,  Yao~Wang$^1$, Alp~Aygar$^1$,  Sohae Chung$^2$, Xiuyuan Wang$^2$, \\ Yvonne W. Lui$^2$, Els Fieremans$^2$, Steven Flanagan$^3$, Joseph Rath$^3$  \\ $^1$Electrical and Computer Engineering Department, New York University \\ $^2$Department of Radiology, New York University \\ $^3$Department of Rehabilitation Medicine, New York University}

\maketitle

\begin{abstract}
In this work, we propose bag of adversarial features (BAF) for identifying mild traumatic brain injury (MTBI) patients from their diffusion magnetic resonance images (MRI) (obtained within one month of injury) by incorporating unsupervised feature learning techniques.
MTBI is a growing public health problem with an estimated incidence of over 1.7 million people annually in US. 
Diagnosis is based on clinical history and symptoms, and accurate, concrete measures of injury are lacking.
Unlike most of previous works, which use hand-crafted features extracted from different parts of brain for MTBI classification, we employ feature learning algorithms to learn more discriminative representation for this task.
A major challenge in this field thus far is the relatively small number of subjects available for training.
This makes it difficult to use an end-to-end convolutional neural network to directly classify a subject from MR images.
To overcome this challenge, we first apply an adversarial auto-encoder (with convolutional structure) to learn patch-level features, from overlapping image patches extracted from different brain regions.
We then aggregate these features through a bag-of-word approach.
We perform an extensive experimental study on a dataset of 227 subjects (including 109 MTBI patients, and 118 age and sex matched healthy  controls), and compare the bag-of-deep-features with several previous approaches.
Our experimental results show that the BAF significantly outperforms earlier works relying on the mean values of MR metrics in selected brain regions.
\end{abstract}

\IEEEpeerreviewmaketitle

\section{Introduction}
Mild traumatic brain injury (MTBI) is a significant  public health problem, which can lead to a variety of problems including persistent headache, memory and attention deficits, as well as behavioral  symptoms.
There is public concern not only regarding civilian head trauma, but sport-related, and military-related brain injuries \cite{mtbi1}.
Up to 20-30\% of patients with MTBI develop persistent symptoms months to years after the initial injury \cite{sohae1}-\cite{sohae2}.
Good qualitative methods to detect MTBI are needed for early triage of patients, believed to improve outcome \cite{sohae3}.


The exploration of non-invasive methodologies for the detection of brain injury using diffusion MRI is extremely promising in the study of MTBI: for example, diffusion tensor imaging (DTI) shows areas of abnormal fractional anisotropy (FA) \cite{ingles}-\cite{kraus} and mean diffusivity (MD) \cite{shenton}  in white matter (WM); and diffusion kurtosis imaging (DKI) shows altered mean kurtosis (MK) within the thalamus in MTBI \cite{grossman}-\cite{stokum}. 
In addition to these conventional measurements, more recently, compartment-specific white matter tract integrity (WMTI) metrics \cite{els} derived from multi-shell diffusion imaging have been proposed to describe microstructural characteristics in intra- and extra-axonal environments of WM, showing reduced intra-axonal diffusivity (Daxon) in the splenium in MTBI \cite{sohae4}. 
Recently, there are new approaches incorporating machine learning based on MR images for MTBI identification  \cite{yuanyi}-\cite{minaee2}.
In spite of the encouraging results, the features used in those works are mainly hand-crafted and may not be the most discriminative features for this task (e.g., mean). 

In this work, we propose a machine learning framework to classify MTBI patients and controls using features extracted from diffusion MRI, particularly in the thalamus, and the splenium of the corpus callosum (sCC), two areas that have been highly implicated in this disorder based on previous works \cite{thalcc1}-\cite{thalcc2}.
The main challenge for using a machine learning approach is that, as many other medical image analysis tasks, we have a relatively small (in machine learning sense) dataset of 227 subjects, and each sample has a very high dimensional raw representation (multiple 3D volumes). 
Therefore, it is not feasible  to directly train a classification network on such datasets.
To overcome this issue, we propose to learn features from local patches extracted from thalamic and splenial regions-of-interest (ROIs) using a deep adversarial auto-encoder to learn patch level features in an unsupervised fashion, and then aggregate the features from different patches through a bag of word representation.
Finally,  feature selection followed by a classification algorithm is performed to identify MTBI patients.
The block diagram of the overall algorithm is shown in Fig. 1.

\begin{figure}[h]
\begin{center}
    \includegraphics [scale=0.48] {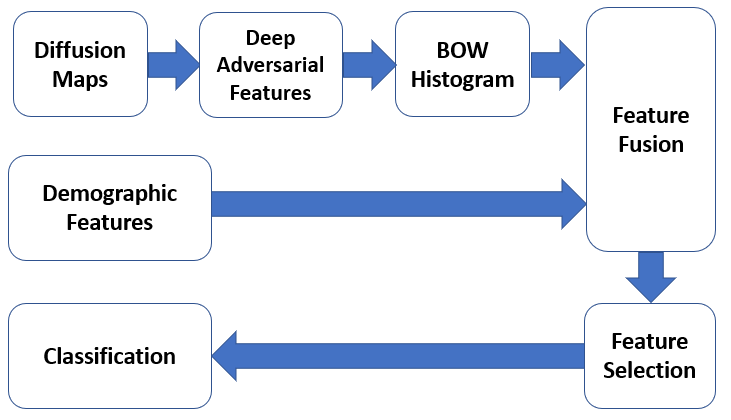}
\end{center}
  \caption{The block-diagram of the proposed MTBI identification algorithm}
\end{figure}

This approach provides a powerful scheme to learn a global representation by aggregating deep features from local regions, and may be a useful approach in cases where there may be a limited number of samples but high dimensional input data (e.g. MRI) can be encountered in specific cases of medical image analysis.
The remaining parts of the paper are organized as follows. Section II provides an overview of previous works on MTBI detection using MR imaging. Section III describes the details of the proposed framework. The experimental studies and comparison are provided in Section IV. Summary and conclusion are stated in Section V.


\section{Previous Works}
Diffusion MRI is one of the most promising imaging techniques to detect in vivo injury in patients with MTBI.
While many of the studies performed over the past decade show group differences between control subjects and MTBI, the consensus report from the American College of Radiology in 2016, cited the utility of these techniques applied to individual subjects as remaining limited \cite{acrm}.

A small number of studies have used machine learning frameworks applied to imaging to identify patients with MTBI.



Lui et al \cite{yuanyi}  proposed a machine learning framework based on 15 features, including 2 general demographic features, 3 global brain volumetric features, and 10 regional brain MRI metrics based on previously demonstrated differences between MTBI and control cohorts.
Mean value of various metrics in different regions are used as the imaging features in their work.
They used Minimum Redundancy and Maximum Relevance (MRMR) for feature selection, followed by a classification algorithm to identify the patients.
They evaluated their model on a dataset of 48 subjects, and showed they are able to identify the MTBI patients with reasonably good accuracy using cross-validation sense.

In \cite{vergara}, Vergara et al investigated the use of resting  state  functional  network  connectivity (rsFNC) for MTBI identification, and did a comparison with diffusion MRI results on the same cohort.
Features based on rsFNC were obtained through group independent component analysis and correlation between pairs of resting state networks.
Features from  diffusion  MRI were obtained using all voxels, the
enhanced Z-score micro-structural assessment for pathology, and the distribution corrected Z-score.
Linear support vector machine \cite{svm} was used for classification and leave-one-out cross validation was used to validate the performance.
They achieved a classification accuracy of 84.1\% with rsFNC features,  compared to 75.5\% with diffusion-MRI features, and 74.5\% using both rsFNC and diffusion-MRI features.

In addition Mitra \cite{mitra} proposed an approach for identifying MTBI based on  FA-based altered structural connectivity patterns derived through the network based statistical analysis of structural connectomes generated from TBI and age-matched control groups. 
Higher order diffusion models were used to map white matter connections between 116 cortical and subcortical regions in this work.
Then they performed network-based statistical analysis of the connectivity matrices to identify the network differences between a representative subset of the two groups. 
They evaluated the performance of their model on a dataset of 179 TBI patients and 146 controls participants, and were able to obtain a mean classification accuracy of 68.16\%$\pm$1.81\%
for classifying the TBI patients evaluated on the subset of the participants that was not used for the statistical analysis, in a 10-fold cross-validation framework.

Also in a previous work \cite{minaee1}, we proposed a machine learning framework for identifying MTBI patients from diffusion MRI, using imaging features extracted from local brain regions, followed by feature selection and classification. 
We evaluated the performance of our model on a dataset of 114 subjects, and showed promising results.

There are several other works using imaging features for MTBI identification. 
For a detailed explanations we refer the readers to \cite{prev1}-\cite{prev4}.

While such previous works show promising results to identify MTBI based on various machine learning approaches, they are limited by the size of training data, and the usefulness of the hand-engineered features used in those works.
To overcome these limitations, in this study we adapted a deep adversarial auto-encoder to learn patch level features in an unsupervised fashion and aggregated the features through a bag of word representation. 
In addition, we included some newly proposed diffusion features to be better able to identify MTBI.

\section{The Proposed Framework}
In this work we propose a machine learning framework for MTBI identification, which relies on the imaging and demographics features.
We derive the imaging features from multi-shell diffusion MR imaging, and bicompartment modeling based on white matter tract integrity (WMTI) metrics derived from diffusion kurtosis imaging (DKI) \cite{els}, that are shown to be promising for assessment of MTBI patients against controls \cite{lui3}-\cite{lui5}.
These metrics are summarized in Table I.
\begin{table}[h]
\centering
\caption{MRI metrics description}
\begin{tabular}{| m{2cm} | m{5.5cm} |}
\hline
MRI Metrics & Metric Description  \\
\hline
\ \ \ \ \ \ AWF & Axonal Water Fraction  \\
\hline
\ \ \ \ \ \ DA & Diffusivity within Axons  \\
\hline
\ \ \ \ \ \ De-par & Diffusion parallel to the axonal tracts in the extra-axonal  \\
\hline
\ \ \ \ \ \ De-perp & Diffusion perpendicular to the axonal tracts in the extra-axonal  \\
\hline
\ \ \ \ \ \ FA & Fractional Anisotropy  \\
\hline
\ \ \ \ \ \ MD & Mean Diffusion  \\
\hline
\ \ AK,\ MK,\ RK & Axial Kurtosis, Mean Kurtosis, Radial Kurtosis  \\
\hline
\end{tabular}
  \label{tab:1}
\end{table}

For imaging features, it is not clear what is the best feature set to use.
Many of the previous works used hand-crafted features for image representation, but since it is not clear beforehand which imaging features are the best for MTBI identification, we propose to learn the feature representation from MR images using a deep learning framework.

Because of the limitation of the number of samples, it is not possible to train a deep convolutional network to directly classify the an entire brain image volume. 
To tackle this problem, and also based on the assumption that MTBI may impact only certain regions in the brain, we propose to represent each brain region by a bag of words (BoW) representation, which is the histogram of different representative patch-level patterns.
By looking at 16x16 patches from thalamus and sCC we get around 454 patches from each subject, which results in more than 100k image patches.
Since we cannot infer patch level labels from subject label, we should use unsupervised feature learning schemes.
We use a recent kind of auto-encoder models, called adversarial auto-encoder, to learn discriminative patch level representations.
The detail of feature extraction and bag of word approach are explained in Section III.A and III.B.


\subsection{Adversarial Auto-Encoder for Patch Feature Learning}
There have been a lot of studies in image processing and computer vision to design features for various applications. 
For patch level description, various  "hand-crafted" features have been developed, such as scale invariant feature transform (SIFT), histogram of oriented gradients (HOG), and local binary pattern (LBP) \cite{sift}-\cite{lbp}.
Although these features perform well for some applications, there are not the best we could do in many cases. 
To derive a (more) optimum set of feature for any task, one can use machine learning techniques to learn the representation. 
Convolutional neural networks are one of the most successful models used for image classification and analysis that jointly learn features and perform classification, and have been used for a wide range of applications from image classification and segmentation, to automatic image captioning \cite{cnn1}-\cite{cnn2}.

The challenge in our problem is that we do not have patch level classification labels, as we cannot assume all patches from a MTBI subject will be "abnormal". 
In order to learn patch-level features without having labels, we employ adversarial auto-encoder \cite{AAE}, an unsupervised feature learning approach.
Adversarial auto-encoder is similar to the regular auto-encoder, in that they both receive an image as the input and perform multiple ``convolution+nonlinearity+downsampling'' layers to encode the image into some latent features, and then use these features to reconstruct the original image through deconvolution.
By doing so, the network is forced to learn some representative information that is sufficient to recover the original image. 
The overall architecture of a regular auto-encoder is shown in Figure 2.
It can be seen that the network consists of two main parts, an encoder and a decoder  \cite{cae1}-\cite{cae2}.
In our work we apply the auto-encoder at the patch level.
After training this model, the latent representation in the mid-layer is used as patch feature representation.
\begin{figure}[h]
\begin{center}
    \includegraphics [scale=0.36] {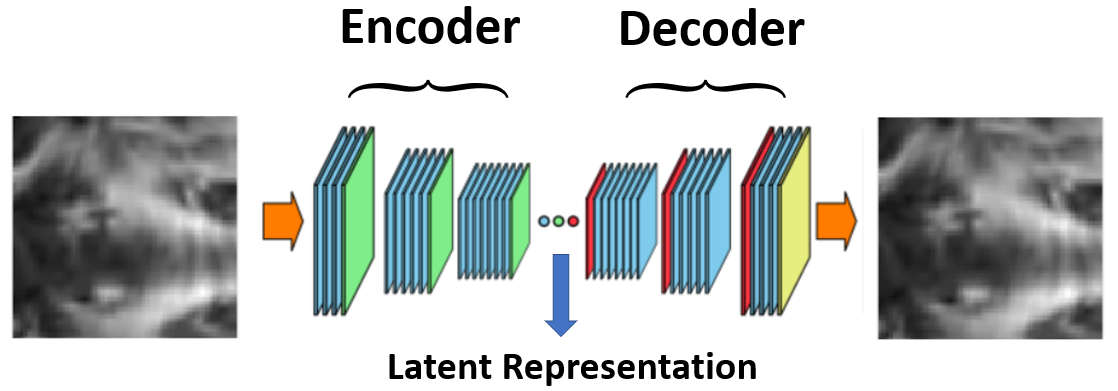}
\end{center}
  \caption{The block-diagram of an example convolutional auto-encoder}
\end{figure}

Adversarial auto-encoder has one more component, by which it enforces some prior distribution on the latent representation. 
As a result, the decoder of the adversarial autoencoder learns a generative model which maps the imposed prior to the data distribution.
The block diagram of an adversarial auto-encoder is shown in Fig 3.
\begin{figure}[h]
\begin{center}
    \includegraphics [scale=0.55] {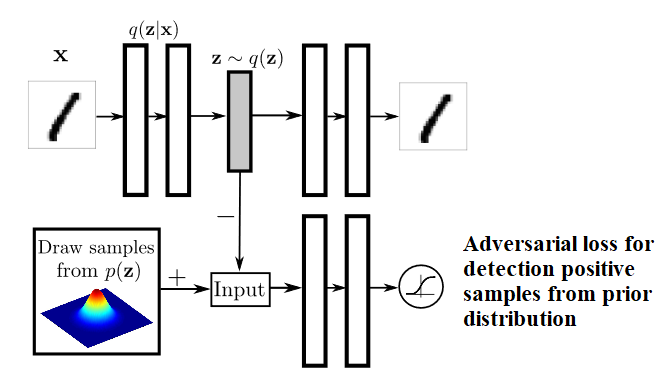}
\end{center}
  \caption{Block-diagram adversarial auto-encoder, courtesy of Makhzani \cite{AAE}}
\end{figure}

As we can see, there is an discriminator network which classifies whether the latent representation of a given sample comes from a prior distribution (Gaussian in our work) or not.
Adding this adversarial regularization guides the auto-encoder to generate latent features with a target  distribution. 

To train the adversarial auto-encoder, we minimize the  loss function in Eq (1) over the training samples.
Note that this loss function consists of two terms, one term describes the reconstruction error, and another one the discriminator loss.
We use the mean square error for the reconstruction loss, and binary cross entropy for the adversarial loss.
The parameter $\lambda$ is a scalar which determines the relative importance of these two terms, and can be tuned over a validation set.
\begin{gather}
 \mathcal{L}_{AAE}= \mathcal{L}_{Rec}+ \lambda \mathcal{L}_{Adv}
\end{gather}

In our study, we train one auto-encoder model for each metric (such as FA, MK, RK, etc.).
Therefore we have multiple networks, where each one  extracts the features from a specific metric (and both regions).
We will provide the comparison between using adversarial auto-encoder for feature learning, with using convolutional auto-encoder, and also with directly using raw voxel values.

\subsection{Bag of Visual Words}
Once the features are extracted from each patch, 
we need to aggregate the patch-level features into a global representation for an entire brain volume.
One simple way could be to get the average representation of patch features as the overall feature.
But this simple approach can lose a lot of information. 
Instead, we use the bag of words (BoW) representation  \cite{bow1} to describe each brain region, which calculates the histogram of representative patterns (or visual words) over all patches in this region.
Bag of visual words is a popular approach in computer vision, and is used for various applications \cite{bow2}-\cite{bow3}.
The idea of bag of visual words in computer vision is inspired by bag of word representation in text analysis, where a document is represented as a histogram of words, and those histograms are used to analyze the text documents. 
Since there is no intrinsic words defined for images, we need to first create the visual words.
To find the visual words, we can apply a clustering algorithm (e.g. k-means clustering) to the patch features obtained from all training patches. 
Given the MR images of a subject, we extract overlapping patches from two designated brain regions (thalamus and sCC).
We then describe a brain region as a histogram of different visual words among all patches in this region. 
The block diagram of the BoW approach is shown in Figure 4.
\begin{figure}[h]
\begin{center}
    \includegraphics [scale=0.3] {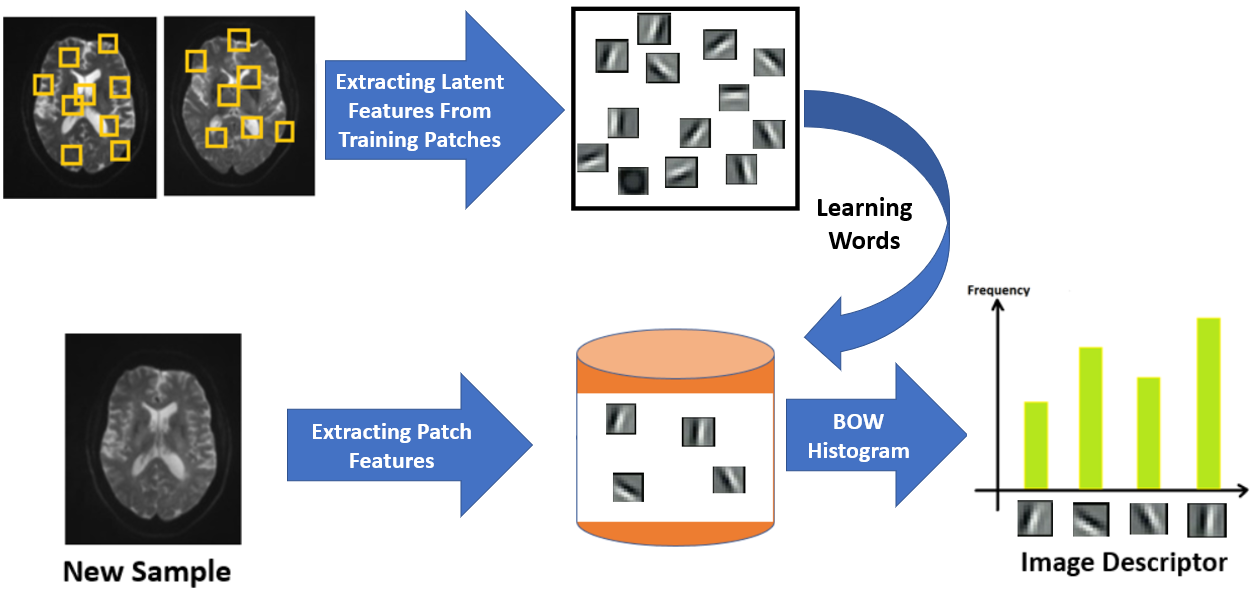}
\end{center}
  \caption{The block-diagram of the proposed BoW approach}
\end{figure}


\subsection{Feature Selection and Classification}
After deriving adversarial features for patches from diffusion MR images, we will get a feature vector per metric and region.
We concatenate the features from different metrics and regions, with demographic features, to form the final feature vector.
We then perform feature selection to minimize the risk of over-fitting before classification \cite{fs1}-\cite{fs2}.
We tried multiple feature selection approaches such as greedy forward selection, max-relevance and min-redundancy (MRMR) \cite{mrmr} and maximum correlation, and it turns out that the greedy forward feature selection works best for our problem.
This approach selects the best features one at a time  with a given classifier, through a cross-validation approach. 
Assuming $S_k$ denotes the best subset of features of size $k$, the $(k+1)$-th feature is selected as the one which results in the highest cross-validation accuracy rate along with the features already chosen (in $S_k$).
One can stop adding features, either by setting a maximum size for the feature set, or when adding more features does not increase the accuracy rate.
For classification, we tried different classifiers (such as SVM, neural network, and random forest) and SVM performed slightly better than the others, and therefore is used in this work.

\section{Experimental Results}
We evaluate the performance of the proposed approach on our dataset of 227 subjects. 
This dataset contains 109 MTBI subjects between 18 and 64 years old, within 1 month of MTBI as defined by the American College of Rehabilitation Medicine (ACRM) criteria for head injury, and 118 healthy age and sex-matched controls. 
The study is performed under institutional review board (IRB) compliance for human subjects research.
Imaging was performed on a 3.0 Tesla Siemens Tim Trio and Skyra scanners including multi-shell diffusion MRI at b-values of 1000 and 2000 s/mm2 at isotropic 2.5mm image resolution.

In-house image processing software developed in MATLAB R2017b was used to calculate 11 diffusion maps including DTI, DKI and WMTI metrics. All diffusion maps in subject space are registered to the Montreal Neurological Institute (MNI) standard template space, by using each subject’s fractional anisotropy (FA) image. 
The regions of thalamus and sCC were extracted from the MNI template and were modified if needed.

We applied BoW approach on three sets of patch-level features, raw voxel values, features generated by a trained convolutional auto-encoder, and features generated by adversarial auto-encoder.
The statistical features from each region for each MR metric consists of 5 different statistics including mean, standard deviation, third and fourth moments, and finally entropy of voxel values in that region.

For the BoW approach, we extract overlapping patches of 16x16 (with stride of 3) from thalamus and sCC, resulting in a total of 454 patches for each subject. 
Therefore in total we get around 103k patches.
Some of the sample patches of 16x16 from various metrics are shown in Figure 5.
\begin{figure}[h]
\begin{center}
    \includegraphics [scale=0.48] {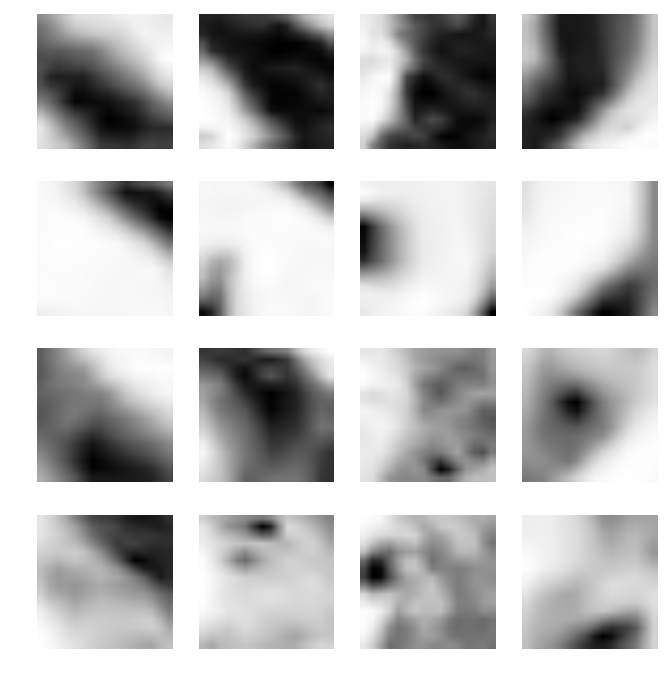}
\end{center}
  \caption{The patches in the first, second, third and fourth rows denote some of the sample patches from FA, MD, Depar and MK metrics.}
\end{figure}

For the convolutional autoencoder, 
the encoder and decoder each have 4 layers, and the kernel size is always set to (3,3). 
The latent feature dimension is 32 for the networks which are trained on individual metrics. 
To train the model, we use one third of all patches, which is around 34k samples. 
The batch size is set to 500, and the model is trained for 10 epochs. 
We use ADAM optimizer to optimize the cost function, with a  learning rate of 0.0003.
The learnt auto-encoder is then used to generate latent features on each overlapping patch in the training images. 
The resulting features are further clustered to $N$ words using K-means clustering. $N$ was varied among 20, 30, and 40. 
Each MR metric in each region is represented by a histogram of dimension $N$.
We used Tensorflow package to train the convolutional-autoencoder model. 

For the adversarial auto-encoder, both encoder and decoder networks contain 4 layers (2 ''convolution+nonlinearity+pooling'' and 2 fully-connected layers), and the discriminator network contains three fully connected layers, to predict whether a latent representation is coming from a prior distribution or not.
The dimension of the latent representation is set to 32 in this case, and the prior distribution of the latent samples is set to be Gaussian.
The learning rate during the update of generative and discriminative networks are set to 0.0006 and 0.0008 respectively.
Pytorch is used to train the adversarial auto-encoder.

For SVM, we use radial basis function (RBF) kernel. 
The hyper-parameters of SVM model (kernel width gamma, and the mis-classification penalty weight, C) are tuned based on a validation set of 45 subjects.
It is worth to mention that, we normalize all features before feeding as the input to SVM, by making them zero-mean and unit-variance.
The SVM module in Scikit-learn package in Python is used to implement SVM algorithm.


In the first experiment, we compare the performance of the proposed bag-of-adversarial-features with global statistical features, BoW feature derived from convolutional auto-encoder \cite{embc2018}, and BoW derived from raw voxel values \cite{minaee1}.

In each case (except for the case with statistical features), a histogram of 20-dimensional is derived for each metric in each of thalamus and sCC regions. 
Then these histograms are concatenated to form the initial image feature, resulting in a 280 dimensional vector, given that there are 9 MR metrics (AWF, DA, De\_par, De\_perp, FA, MD, AK, MK, RK) in sCC, and 5 MR metrics in thalamus (FA, MD, AK, MK, RK).
Together with additional 2 demographic features (age and sex), the total feature dimension is 282.

To perform feature selection and evaluate the model performance, we use a cross validation approach, where each time we randomly take 20\% of the samples for validation, and the rest for training.
We repeat this procedure 50 times (to decrease sampling bias), and report the average validation error as the model performance.

To have a better generalization accuracy analysis, 
once the features are chosen, we divide the dataset into three sets, training, validation, and heldout samples, where we train the model on the training set and find the optimum values of the SVM hyper-parameters using the validation set, and evaluate the model performance on the heldout set.
In each run, we randomly pick 45 samples out of the entire 227 samples as the heldout set.
We then run cross validations 50 times within the remaining data (using 137 samples for training and 45 samples for validation), to generate 50 models, and use the ensemble of 50 models to make prediction on the held-out set and calculate the classification accuracy. 
We repeat this 4 times, each time with a different set of 45 heldout samples chosen randomly and report the average accuracy. 
The average accuracy for the validation and heldout sets for four different approaches are given in Table II.
The results reported here use first 10 chosen features for each method.
As we can see, adversarial-BoW approach achieves higher accuracy  over previously used features.
One reason that adversarial features are better than the convolutional features could be that by regularizing the latent representations to be drawn from a prior distribution, it is much easier for the network to converge.
Interestingly enough, the heldout accuracies are close to validation accuracies, which could be a good indicator of the generalizability of the proposed features.

\begin{table}[ht]
\centering
  \caption{Performance comparison of different approaches}
  \centering
\begin{tabular}{|m{3.1cm}|m{2.2cm}|m{2.2cm}|}
\hline
The Algorithm  & Classification Rate on Validation Set & Classification Rate on Heldout Set\\
\hline
The selected subset of \ \ \ \ statistical features \cite{minaee1}  &  \ \ \  \ \ \  \ \ 78\% & \ \ \  \ \ \  \ 76.6\% \\
\hline
BoW on raw patches  with 20D histograms \cite{minaee1}  &  \ \ \ \ \  \ \   \ 80.9\% & \ \ \  \ \ \  \ 79.9\%\\
\hline
The Convlutional-BoW with 20D histograms \cite{embc2018}  &  \ \  \  \ \ \ \ \ 81.2\% & \ \ \  \ \ \  \ 79.9\%\\
\hline
The proposed Adversarial-BoW (20D histograms)  &  \ \  \ \  \ \ \ \ 84.2\% & \ \ \  \ \ \  \ 83.8\%\\
\hline
\end{tabular}
\label{TblComp}
\end{table}

To evaluate the robustness of the model predictions for heldout samples, we evaluated the standard deviation of prediction accuracy over these 50 ensemble predictors. The standard deviations for different feature sets are shown in Table III.
As we can see from this table, the prediction accuracy has a small variation among different models, which is a good indicator of good generalization.
\begin{table}[ht]
\centering
  \caption{Analysis of mean and standard deviation of heldout accuracy of different approaches}
  \centering
\begin{tabular}{|m{3.1cm}|m{2.2cm}|m{2.2cm}|}
\hline
The Algorithm  & Classification Rate Mean & Classification Rate Standard Deviation\\
\hline
The selected subset of \ \ \ \ statistical features \cite{minaee1}  &  \ \ \  \ \ \  \ \ 76.6\% & \ \ \  \ \ \  \ 3.05\% \\
\hline
BoW on raw patches  with 20D histograms \cite{minaee1}  &  \ \ \ \ \  \ \   \ 79.9\% & \ \ \  \ \ \  \ 3.67\%\\
\hline
The Convlutional-BoW with 20D histograms \cite{embc2018}  &  \ \  \  \ \ \ \ \ 79.9\% & \ \ \  \ \ \  \ 3.08\%\\
\hline
The proposed Adversarial-BoW (20D histograms)  &  \ \  \ \  \ \ \ \ 83.8\% & \ \ \  \ \ \  \ 2.85\%\\
\hline
\end{tabular}
\label{TblComp}
\end{table}

With forward feature selection using the SVM classifier,  the optimal feature subsets  (with maximum feature number set at 10) chosen from different features extraction algorithms are listed in the Table IV.
It is worth mentioning that the chosen features vary between techniques and it is not entirely clear why. Of note, our Adversarial BoW technique selects adversarial visual words from kurtosis measures of the thalamus and axonal diffusion, DA, of the splenium of the corpus callosum, both measures which have previously been implicated in differentiating MTBI patients from controls \cite{sohae4}, \cite{fea1}-\cite{fea2}.

\begin{table}[ht]
\centering
  \caption{Performance comparison of different approaches. Note that Thal refers to the thalamus region, and sCC refers to Splenium subregion within Corpus Callosum.}
  \centering
\begin{tabular}{|m{2.2cm}|m{5.3cm}|}
\hline
The Algorithm  & Chosen Features' Metric and Region\\
\hline
Statistical features  &  MD in sCC (mean), FA in sCC (entropy), AK in sCC (mean), FA in Thal (mean), MD in Thal (var), Depar in sCC (entropy), MK in sCC (entropy), AWF in sCC (mean), Deperp in sCC (entropy), MK in Thal (entropy)\\
\hline
Raw-BoW   &  FA in sCC, MD in Thal, MK in sCC, AK in Thal, MD in sCC, AWF in sCC, AK in Thal, Depar in sCC, AK in sCC, FA in Thal\\
\hline
Conv-BoW  &  FA in Thal, AK in Thal, Depar in sCC, MK in sCC, RK in sCC, AK in sCC, MD in Thal, RK in sCC, MD in Thal, MD in sCC\\
\hline
Adversarial-BoW  &  MD in Thal, AK in Thal, RK in sCC, FA in Thal, Deperp in sCC, MD in Thal, DA in sCC, MD in sCC, Deperp in sCC, Depar in sCC\\
\hline
\end{tabular}
\label{TblComp}
\end{table}

Besides classification accuracy, we also report the sensitivity and specificity, which are important in the study of medical data analysis.
The sensitivity and specificity are defined as in Eq (2), where TP, FP, TN, and FN denote true positive, false positive, true negative, and false negative respectively.
In our evaluation, we treat the MTBI subjects as positive. 
\begin{gather}
 \text{Sensitivity}= \frac{\text{TP}}{\text{TP+FN}} \ , 
\ \ \ \ \text{Specificity}= \frac{\text{TN}}{\text{TN+FP}} 
\end{gather}
The sensitivities, specificities and F1-scores for different features are shown in Table V. 

\begin{table}[ht]
\centering
  \caption{Performance comparison of different approaches}
  \centering
\begin{tabular}{|m{2cm}|m{1.5cm}|m{1.5cm}|}
\hline
The Algorithm   &  Sensitivity & Specificity \\
\hline
Statistical   & \ \ \ \ 82.8 & \ \ \ \ 74.1  \\
\hline
Raw-BoW   & \ \ \ \  79.5 & \ \ \ \  82.3  \\
\hline
Conv-BoW   & \ \ \ \  80.2 & \ \ \ \ 82.1 \\
\hline
Adv-BoW   & \ \ \ \  86.1 & \ \ \ \  81.8   \\
\hline
\end{tabular}
\label{TblComp}
\end{table}

Figure 6 denotes the validation classification accuracies, sensitivities and specificities achieved by different ratios of training samples using adversarial features.
We see that using approximately 80\% of training samples gives reasonably well validation performance, and we do not gain much by using higher ratios of training samples.
Similar trends were observed with other features as well.
All other results reported in this paper were using 80\% samples for training and 20\% for validation, in the cross validation study.
\begin{figure}[h]
\begin{center}
    \includegraphics [scale=0.3] {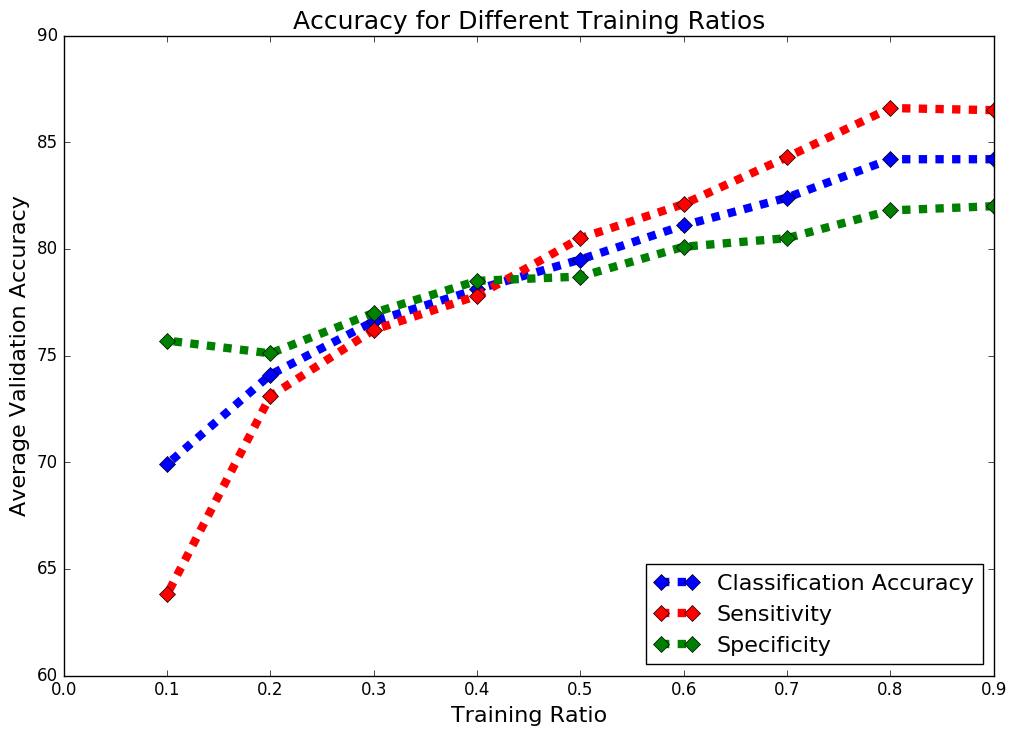}
\end{center}
  \caption{The model performance for different training ratios}
\end{figure}

In Figure 7, we present the receiver operating characteristic (ROC) curve for different set of features on heldout samples.
ROC curve is a plot which illustrates the diagnostic ability of a binary classifier system as its discrimination threshold is varied.
This curve is created by plotting the true positive rate (i.e. sensitivity) against the false positive rate for various threshold settings. 
Recall that we use the mean prediction of 50 classifiers to predict whether a subject has MTBI. Previously reported results are obtained with a threshold of 0.5 on the mean prediction. 
The ROC curve is derived by varying the threshold from 0 to 1 with a stepsize of 0.05.
As we can see the adversarial and convolutional features provide higher  sensitivities  under the same false positive rate than the other two methods, and overall have larger areas under the curve (AUCs).
\begin{figure}[h]
\begin{center}
    \includegraphics [scale=0.25] {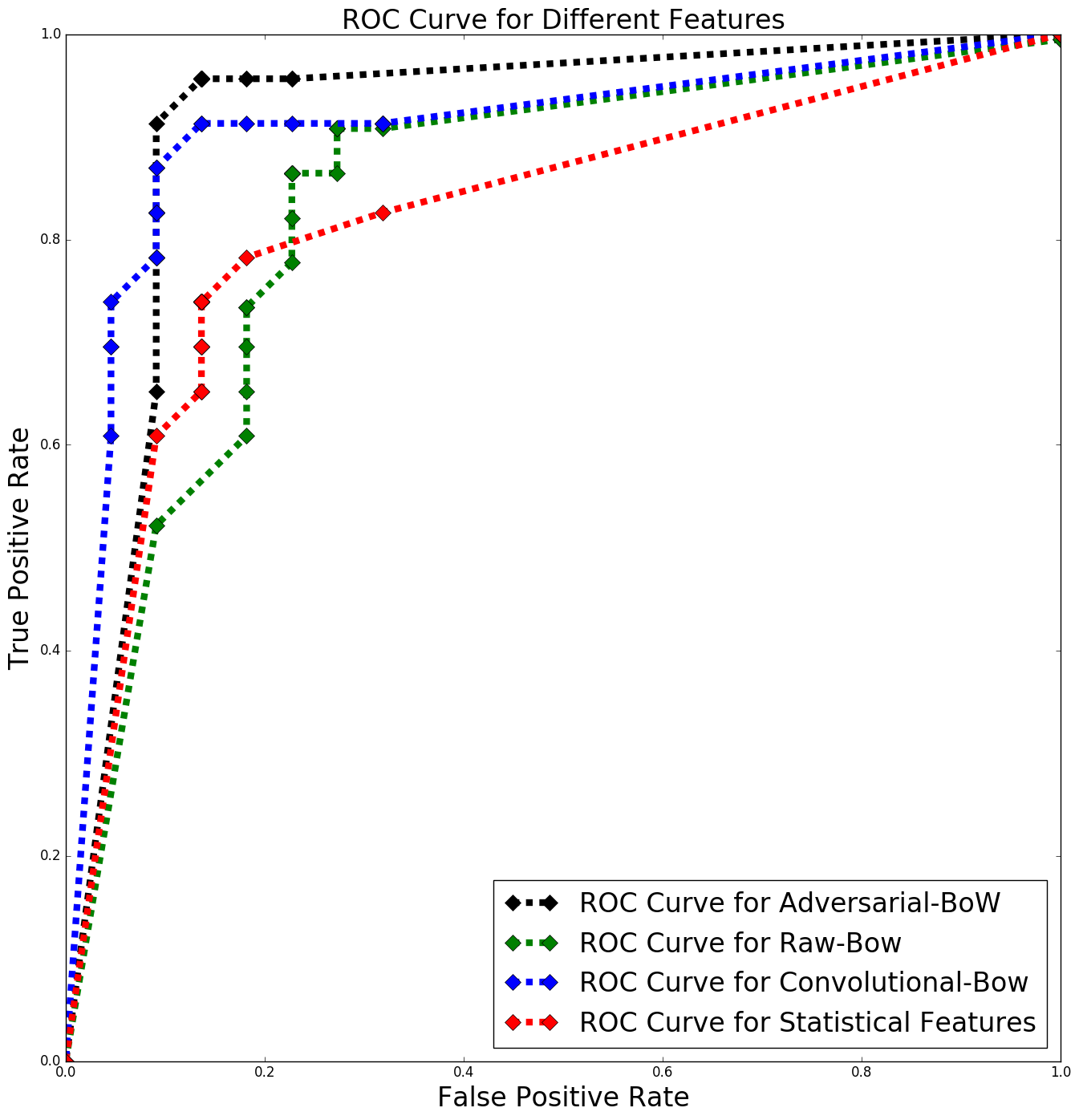}
\end{center}
  \caption{The ROC curve for different features}
\end{figure}

We also studied the classification performance using feature subset of different sizes. 
These results are shown in Figure 8.
We see that with more than 10 features, it is possible to further improve the results slightly, except with the statistical features.
\begin{figure}[h]
\begin{center}
    \includegraphics [scale=0.3] {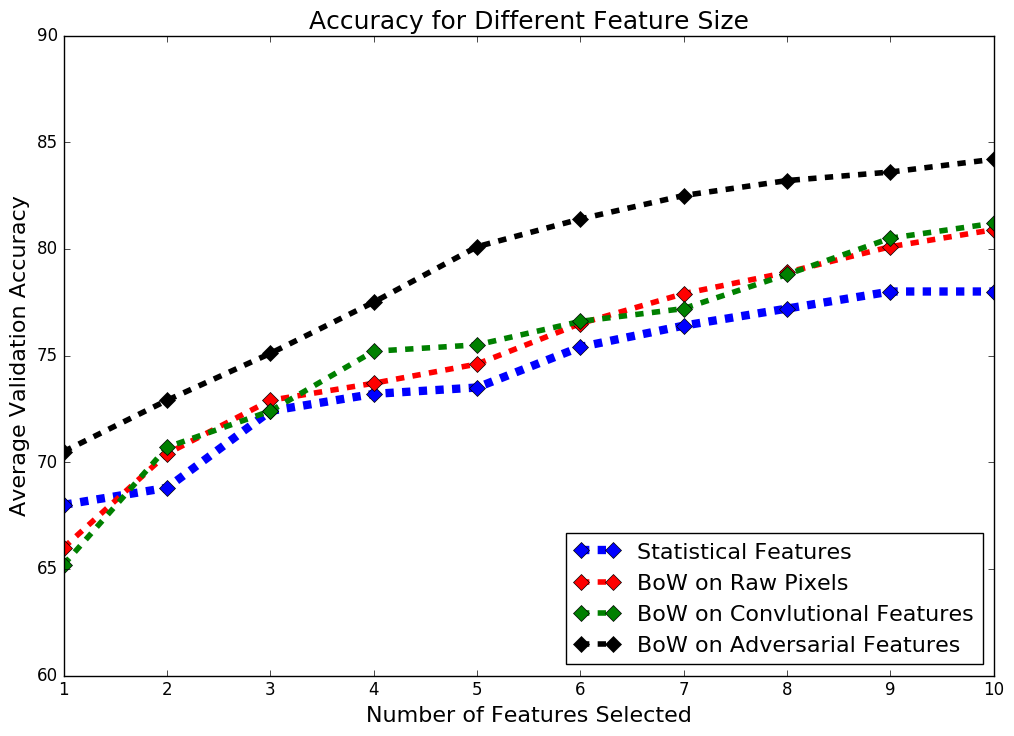}
\end{center}
  \caption{The model performance for feature sets of different sizes}
\end{figure}

In another experiment, we evaluated the impact of histogram dimension on the classification performance.
We generated histograms of 10, 20, and 30 dimensions, respectively, for each metric and region, and performed classification. 
The accuracies are reported in Table VI.
Using 20 dimensional histogram yields the best performance on the held-out set.
\begin{table}[ht]
\centering
  \caption{Performance comparison of different approaches}
  \centering
\begin{tabular}{|m{2.5cm}|m{2.2cm}|m{2.2cm}|}
\hline
The BoW Histogram Dimension  & Classification Rate on Validation Set & Classification Rate on Heldout Set\\
\hline
\ \ \ \ \ \ \ \ \ \ \ 10  &  \ \ \  \ \ \  \ \ 80\% & \ \ \  \ \ \  \ 79.4\% \\
\hline
\ \ \ \ \ \ \ \ \ \ \ 20  &  \ \  \ \  \ \ \ \ 84.2\% & \ \ \  \ \ \  \ 83.8\%\\
\hline
\ \ \ \ \ \ \ \ \ \ \ 30  &  \ \  \ \  \ \ \ \ 84.5\% & \ \ \  \ \ \  \ 82.9\%\\
\hline
\end{tabular}
\label{TblComp}
\end{table}

Finally, we  present the average histograms  of MTBI, and control subjects. 
These histograms and their difference are shown in Fig 9.
As we can see MTBI and control subjects have clear differences in some parts of these representations.
\begin{figure}[h]
\begin{center}
    \includegraphics [scale=0.21] {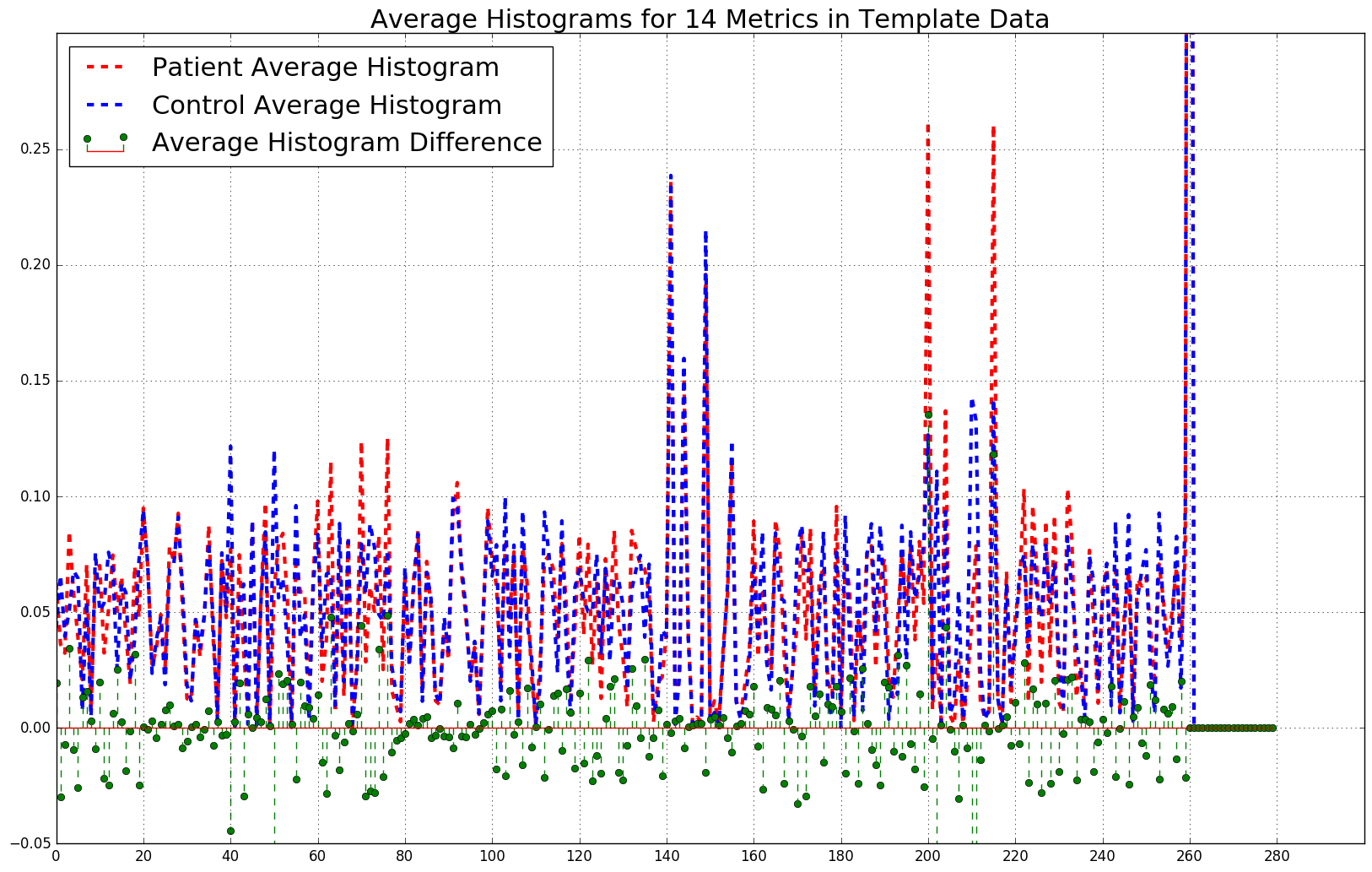}
\end{center}
  \caption{Adversarial-BoW histograms of patients and controls}
\end{figure}

We also find the average histogram over the chosen words for MTBI and control subjects.
These histograms are shown in Fig 10.
As we can, MTBI and control subjects have clear differences over the chosen words.
For example the first two words, are less frequent in patients, than in controls.
\begin{figure}[h]
\begin{center}
    \includegraphics [scale=0.27] {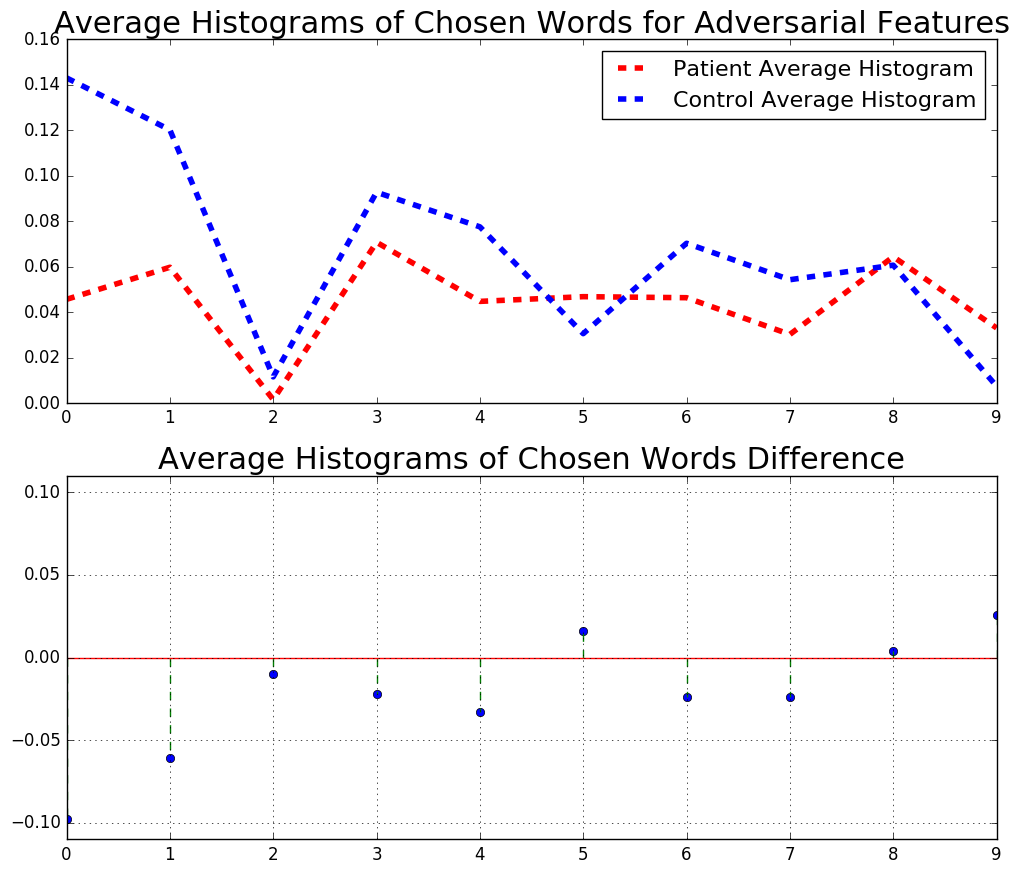}
\end{center}
  \caption{Adversarial-BoW histograms over the chosen words}
\end{figure}

We also tried to localize the chosen words within the brain. To do so, each time we focus on one of the words chosen by the proposed classification algorithm, and then go over all patches of 16x16 in thalamus and sCC (by shifting the patches with some stride) to see if they are quantized to the chosen word. If so, we increment by one the voxel values in that patch to active regions, and repeat this procedure for the remaining patches. 
Here, we provide the heatmaps of two patients and two control subjects for two chosen words. 
These heatmaps are illustrated in Fig 11-12.
As we can see from Fig 11, this word is much more frequent in patient subjects, than in controls.
It could imply that, this word has some MTBI related information. 
Note that, for each case, the top two rows denote the heatmap of a specific word over different slices, and the bottom two rows denote the actual metric of those slices. 
We intentionally increased the contrast of the actual metrics in the bottom to row, for better illustration.

\begin{figure*}[ht]
        \centering
        \vspace{-0.5cm}
        \begin{subfigure}[b]{0.42\textwidth}
                \includegraphics[width=\textwidth]{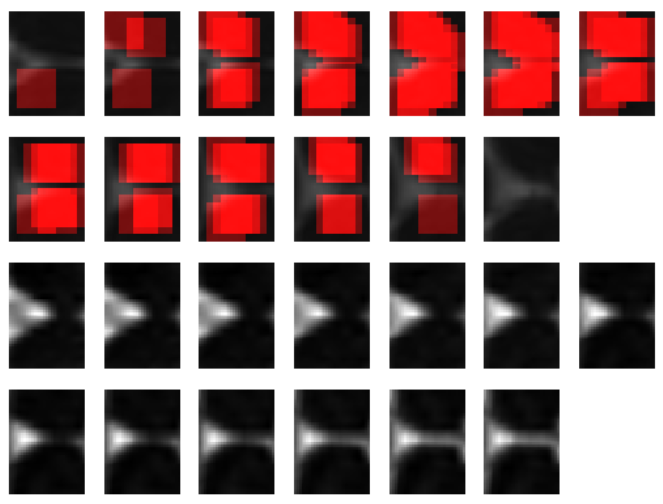}
                                \vspace{-0.03cm}
          \hspace{0.5cm}    
        \end{subfigure}%
        ~ 
        \begin{subfigure}[b]{0.42\textwidth}
       \hspace{1cm}
                \includegraphics[width=\textwidth]{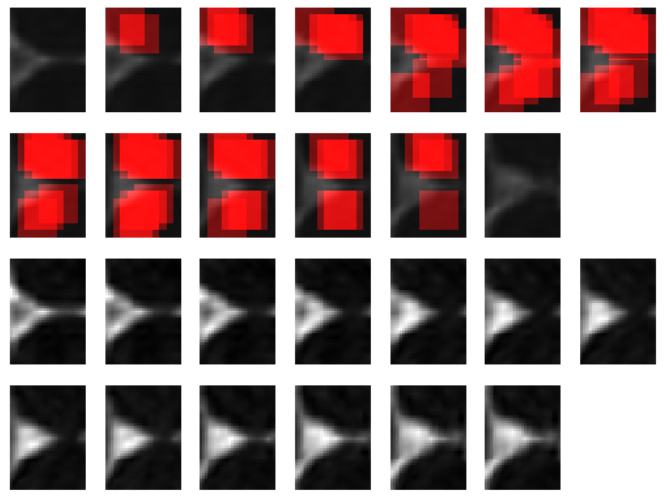}
                \vspace{-0.025cm}
            \hspace{2.02cm} 
        \end{subfigure}
         \\[1ex]
		\begin{subfigure}[b]{0.42\textwidth}
        \vspace{0.5cm}
                \includegraphics[width=\textwidth]{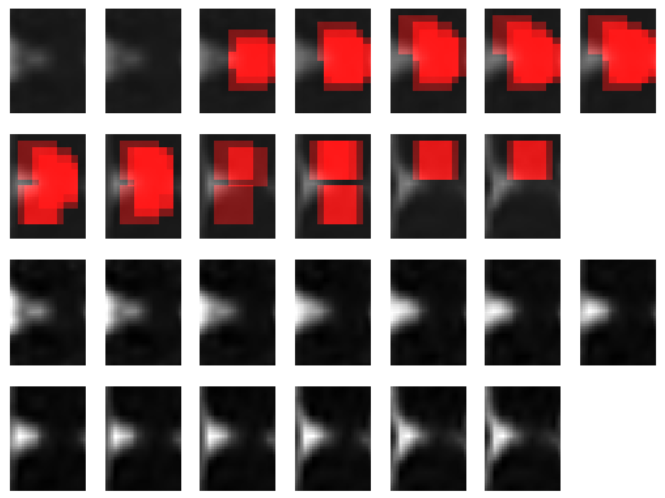}
                \vspace{-0.03cm}
            \hspace{2cm} 
        \end{subfigure}%
        ~ 
        \begin{subfigure}[b]{0.42\textwidth}
        \hspace{1cm}
                \includegraphics[width=\textwidth]{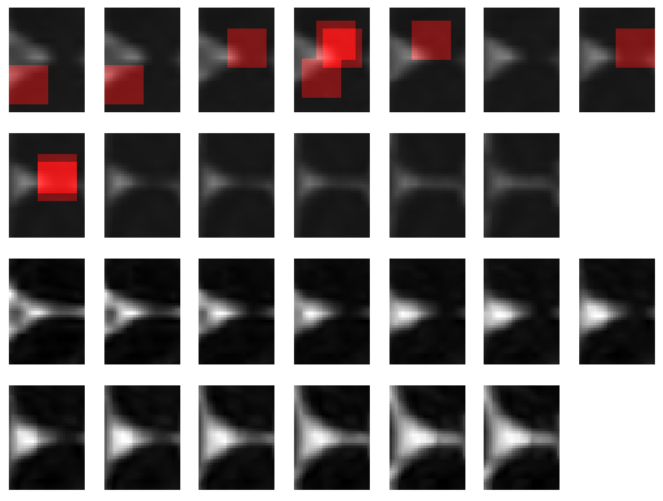}
                \vspace{-0.025cm}
            \hspace{2cm} 
        \end{subfigure}
        \caption{Localization heatmaps corresponding to a chosen word in thalamus and MD metric. The figures in top row denote the heatmaps for two patient subjects, and the heatmaps in bottom row denotes the heatmaps for two controls. In each figure, the first two rows denote the location of chosen words in different parts of 13 thalamus slices, and the next two rows denote the actual MD metrics in thalamus for those subjects.}
\end{figure*}

\begin{figure*}[ht]
        \centering
        \vspace{-0.5cm}
        \begin{subfigure}[b]{0.42\textwidth}
                \includegraphics[width=\textwidth]{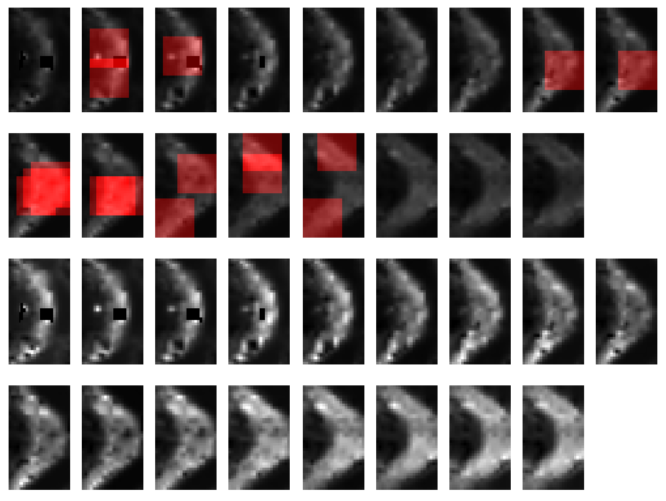}
                                \vspace{-0.03cm}
          \hspace{0.5cm}    
        \end{subfigure}%
        ~ 
        \begin{subfigure}[b]{0.42\textwidth}
       \hspace{1cm}
                \includegraphics[width=\textwidth]{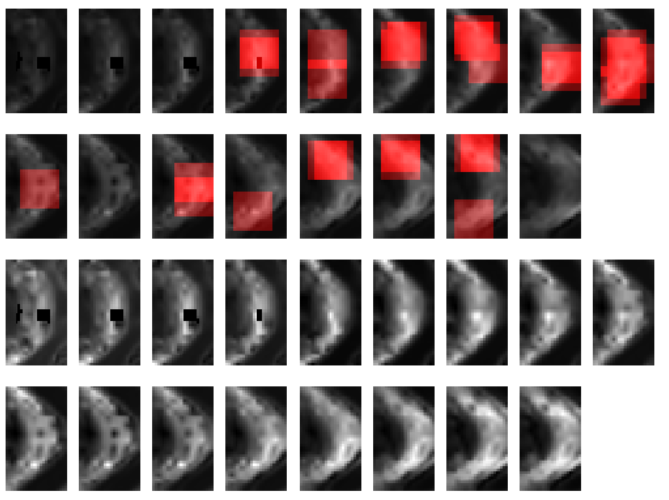}
                \vspace{-0.025cm}
            \hspace{2.02cm} 
        \end{subfigure}
         \\[1ex]
		\begin{subfigure}[b]{0.42\textwidth}
                \includegraphics[width=\textwidth]{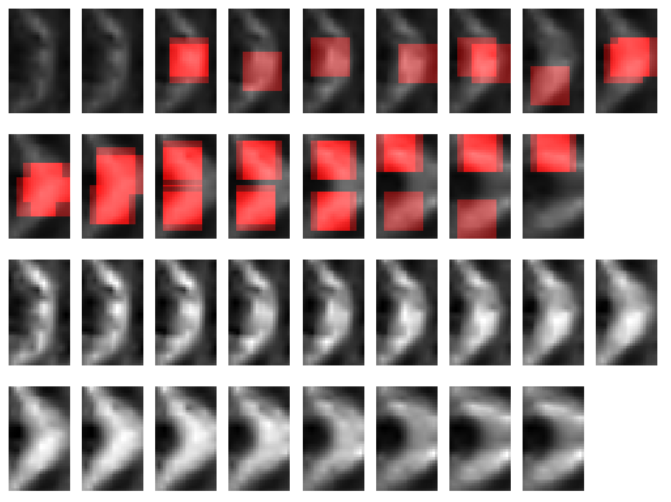}
                \vspace{-0.03cm}
            \hspace{2cm} 
        \end{subfigure}%
        ~ 
        \begin{subfigure}[b]{0.42\textwidth}
        \hspace{1cm}
                \includegraphics[width=\textwidth]{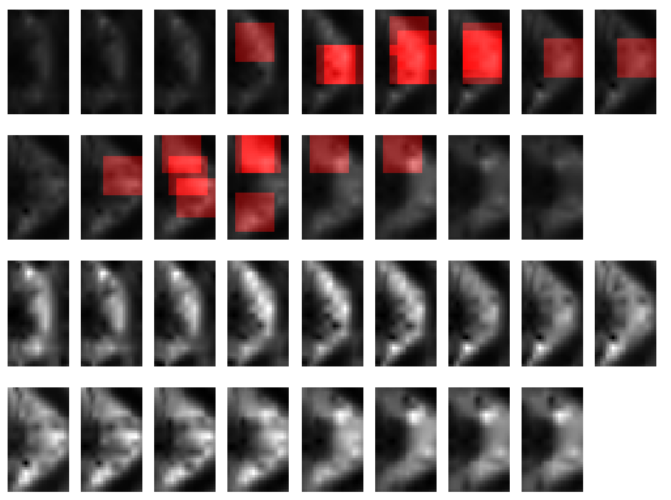}
                \vspace{-0.025cm}
            \hspace{2cm} 
        \end{subfigure}
        \caption{Localization heatmaps corresponding to a chosen word in sCC and RK metric. The figures in top row denote the heatmaps for two patient subjects, and the heatmaps in bottom row denotes the heatmaps for two controls. In each figure, the first two rows denote the location of chosen words in different parts of 13 thalamus slices, and the next two rows denote the actual MD metrics in thalamus for those subjects.}
\end{figure*}

\section{Conclusion}
In this work, we propose an unsupervised learning framework for MTBI identification from diffusion MR images using a dataset of 227 subjects.
We first learn a good representation of each brain regions, by employing a deep unsupervised learning approach that learns feature representation for image patches, followed by aggregating patch level features using bag of word representation to form the overall image feature.
These features are used along with age and gender as the final feature vector.
Then greedy forward feature selection is performed to find the best feature subset, followed by SVM to perform classification.
Through experimental studies, we show that by learning deep visual features at the patch level, we obtain significant gain over using mean values of MR metrics in brain regions.
The performance is also improved over the approach where the visual words are determined based on the raw image patch representation.
Furthermore, we found that the features learnt with an adversarial autoencoder are more powerful than a non-adversarial autoencoder. This methodology may be of particular use for learning features from datasets with relatively small number of samples, as can be encountered in some medical image analysis studies. The learned features could also be used for tasks other than classification such as long-term outcome prediction.

\section*{Acknowledgment}
Research reported in this paper is supported in part by grant funding from the National Institutes of Health (NIH): R01 NS039135-11 and R21 NS090349, National Institute for Neurological Disorders and Stroke (NINDS). This work is also performed under the rubric of the Center for Advanced Imaging Innovation and Research (CAI2R, www.cai2r.net), a NIBIB Biomedical Technology Resource Center (NIH P41 EB017183).
We would like to thank both NIH and CAI2R for supporting this work.
The content is solely the responsibility of the authors and does not necessarily represent the official views of the NIH.

\end{document}